\documentclass[runningheads]{llncs}

\usepackage{booktabs}
\usepackage{array}
\usepackage{graphicx}
\usepackage{pgfplots}
\usepackage{xcolor}
\pgfplotsset{compat=1.18}

\definecolor{linkblue}{rgb}{0.0,0.32,0.68}
\usepackage[breaklinks,colorlinks,citecolor=linkblue,linkcolor=linkblue,urlcolor=linkblue]{hyperref}

\newcommand{\modelgemma}{Gemma 4 12B}
\newcommand{\modelqwen}{Qwen3-VL-8B}
\newcommand{\fmetric}{F1@0.3}
\newcommand{\fstrict}{F1@0.5}

\title{Dense Coordinate-List Fine-Tuning Induces a Controllable Interference Surface in Vision-Language Models}
\titlerunning{Dense Coordinate-List Fine-Tuning}

\author{Chenyu Zhou\inst{1,*} \and Qiliang Jiang\inst{2,*} \and Boguang Pan\inst{3}}
\authorrunning{C. Zhou et al.}
\institute{
School of Engineering, Institute of Science Tokyo, Japan\\
\email{zhou.c.76d6@m.isct.ac.jp}
\and
College of Control Science and Engineering, Zhejiang University, China\\
\email{jiangqiliang@zju.edu.cn}
\and
Graduate School of Information, Production and Systems, Waseda University\\
\email{jxdoudou@suou.waseda.jp}\\[0.5ex]
\textsuperscript{*}These authors contributed equally to this work.
}

\begin{document}
\maketitle

\begin{abstract}
Fine-tuning vision-language models to emit dense coordinate lists improves visual grounding but also changes how models serialize, repeat, and terminate structured outputs. We study this behavior as a generation and control surface. In \modelgemma, high-capacity q/k/v/o LoRA raises class-aware \fmetric\ from 0.007 to 0.448 while inducing repeated-tail pressure (duplicate rate 0.080, max repeat 23). A q/v rank sweep keeps max repeat at 21--22 across ranks 4--64, showing capacity persistence. The target signal is separable: object-level repeat-stop removes exact repeated records (duplicate rate 0.000, max repeat 1) while preserving F1 (0.494 to 0.490) and stricter \fstrict\ (0.381 to 0.385). Structure-axis probes localize the effect to bbox-coordinate object lists; dense non-bbox and spatial/count JSON remain repeat-clean, including under high-capacity adapters. \modelqwen\ reproduces a clean controlled endpoint (\fmetric\ 0.318, duplicate rate 0.000), and COCO 2017 reproduces acquisition plus duplicate pressure. Dense coordinate-list adaptation therefore creates a structure-bound, cross-family interference surface that can be measured and controlled.
\end{abstract}

\section{Introduction}

\begin{figure}[t]
\centering
\includegraphics[width=\textwidth]{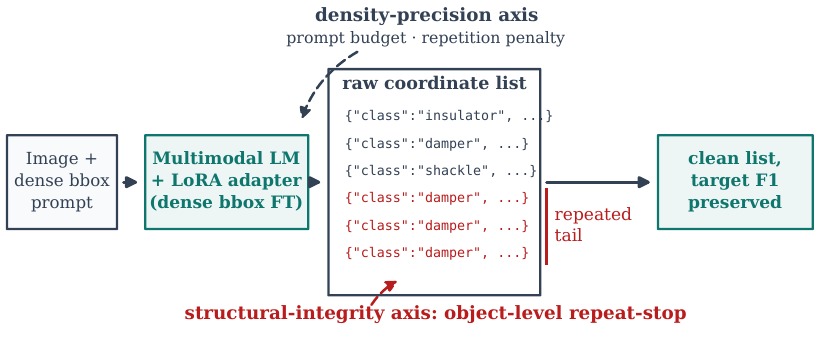}
\caption{The interference surface and its two-axis decomposition. Dense bbox fine-tuning raises target localization and induces a repeated tail on the same generation surface (red records). The density-precision axis (prompt budget, repetition penalty) selects the operating density; the structural-integrity axis (object-level repeat-stop) closes the list at the first exact repeated record, reaching the clean endpoint with target F1 preserved.}
\label{fig:architecture}
\end{figure}

Visual grounding fine-tuning is usually evaluated as target localization. In a generative multimodal language model, dense grounding is also structured-output control. Each object record couples a class label with numeric coordinates, and the full response couples many records with syntactic closure and list termination. Dense bbox generation therefore exposes fine-tuning interference directly at the output surface: adaptation does not only teach the model where objects are---it also reshapes how lists continue and when they stop.

This paper studies that surface directly (Figure~\ref{fig:architecture}). We define \emph{generation surface} as the model's distribution over structured outputs under a given fine-tuning and decoding configuration. We define \emph{control surface} as the measured trade-off among target F1, parse stability, prediction density, duplicate pressure, and termination behavior. Dense coordinate lists exercise all of these axes at once. We organize this surface along two orthogonal control axes: a \emph{density-precision axis} that selects how many objects the model commits to, and a \emph{structural-integrity axis} that keeps the emitted list free of exact repeated records.

Our central result is that this interference is not diffuse capability loss but a localized, decomposable surface. \modelgemma\ supplies the mechanism chain: high-capacity adaptation over q/k/v/o raises target localization and induces repeated-tail pressure on the same surface, while removing exact repeated records preserves target F1. We navigate the surface along these two axes, and we localize the pressure to bbox-coordinate object lists through a structure-axis sweep that leaves dense non-bbox JSON and spatial/count JSON repeat-clean under the rank-8 q/v adapter in both families and under the high-capacity rank-32 q/k/v/o adapter. \modelqwen\ reproduces the mechanism under a different architecture and coordinate protocol, and a COCO 2017 object-centric reproduction shows that acquisition and repeated-list pressure transfer beyond the original industrial dataset. Four exclusions pin down its nature: the pressure is absent from equally dense non-bbox JSON (not a property of structured output in general), survives a sixteen-fold change in adapter capacity (not a capacity artifact), reproduces across model families and on a second public dataset (not an implementation artifact), and carries a negligible share of the localization signal (not load-bearing). What remains is a structure-bound, decomposable interference surface.

\paragraph{Contributions.} We make the following contributions:
\begin{itemize}
\item We frame dense coordinate-list generation as a concentrated stress surface that couples visual localization, numeric serialization, repeated-class regulation, and list termination, and we make this surface measurable through paired target, repetition, density, structural, and structure-axis metrics on a single held-out evaluation split.
\item We show that high-capacity \modelgemma\ adaptation reaches strong target grounding (\fmetric\ 0.448) while exposing repeated-tail pressure on the same surface, and that lightweight inference-time controls separate useful localization from exact repeated records, reaching a clean endpoint at \fmetric\ 0.490 and \fstrict\ 0.385 with duplicate rate 0 and max repeat 1.
\item We isolate the pressure to bbox-coordinate object lists through a structure-axis sweep: dense non-bbox JSON and spatial/count JSON remain repeat-clean in both model families and under the high-capacity adapter, showing that the interference is specific to coordinate-list structure rather than dense JSON generation in general.
\item We validate the mechanism beyond the primary setting: \modelqwen\ reproduces the density/repetition relation and a high-capacity controllable endpoint under a different architecture and coordinate protocol, and COCO 2017 reproduces the acquisition/repetition signature on a public object-centric detection subset.
\end{itemize}

\section{Dense Coordinate-List Interference}

A dense bbox output is a serialized sequence of class--coordinate records. The model must emit class labels, coordinate numbers, separators, object boundaries, and a final closure. Fine-tuning can move target localization and list-generation behavior together, producing a surface rather than a single scalar target metric.

Dense coordinate lists stress a different part of the interface from short VQA, captioning, or sparse single-box grounding. A short answer terminates after one span, whereas a dense coordinate list must hold a parseable schema across many object records while regulating which classes repeat and deciding where the list ends. Visual localization, numeric serialization, repeated-class regulation, and list termination are exercised simultaneously, which makes the surface legible: a single adapter can move target accuracy and list-generation behavior at the same time.

Three structural properties make the coordinate-list regime distinctive among structured outputs, and each corresponds to a measurement in this paper. First, repeated class labels are legitimate within a single output---an image can contain several instances of the same class---so the model cannot rely on class novelty to regulate termination; this is where exact-record repetition concentrates (Table~\ref{tab:structure}). Second, the record payload is numeric: coordinates carry the localization signal, and the same record can recur with identical numbers, which is what the exact-duplicate metric detects and what object-level repeat-stop intercepts. Third, list termination is a learned decision with no schema-forced endpoint, which is why prompt budgets move the operating point so directly (Table~\ref{tab:densitysweep}). Dense non-bbox JSON carries the same schema burden at even greater output length, with fixed cardinality and keys, and it stays repeat-clean under every adapter we test---the contrast that localizes the interference to the coordinate-list structure itself.

\section{Experimental Setup}
\label{sec:setup}

\paragraph{Models.} \modelgemma\ is a unified decoder-only multimodal model of the Gemma family \cite{gemma-report}, in which visual and text tokens share one transformer stack without a separate vision encoder; it is the primary model for exposing the interference surface. \modelqwen\ is an encoder-based multimodal model of the Qwen-VL family \cite{qwen-vl} and serves as a same-data cross-family control. The families use different coordinate protocols: \modelgemma\ is read in pixel coordinates and \modelqwen\ in its native $0$--$1000$ normalized grid. Both families are evaluated on the same evaluation split with identical metrics.

\paragraph{Data.} We fine-tune and evaluate on InsPLAD industrial inspection imagery \cite{insplad}. The adapter is trained on 160 images and evaluated on an 80-image held-out evaluation split with zero filename overlap against the training split. Every reported InsPLAD operating point uses this same split, so target, repetition, density, structural, and structure-axis metrics are directly comparable. For the second-dataset reproduction, we construct a public COCO 2017 \cite{coco} object-centric dense-bbox subset from val2017 detection annotations: the selected split excludes person and car, uses the ten most frequent remaining object classes, keeps images with two to six target objects, and separates 780 training images from a 120-image evaluation split; a scale-matched 160/80 split drawn from the same subset mirrors the InsPLAD protocol. Because every evaluation image has at most six target objects, the COCO reproduction uses a matched prompt budget of six for both base and adapted models.

\paragraph{Adapters.} The main \modelgemma\ chain uses high-capacity q/k/v/o LoRA \cite{lora} with rank 32, $\alpha=64$, and 42.7M trainable parameters. \modelqwen\ uses a rank-32 q/v adapter for its high-capacity controlled endpoint. For the structure-axis analysis we additionally use a capacity-controlled rank-8 q/v adapter, which isolates the effect of output structure under a matched adapter setting. For the capacity sweep, we hold the \modelgemma\ module set fixed at q/v and vary LoRA rank over 4/8/16/32/64.

\paragraph{Surface controls.} We organize control of the interference surface into two orthogonal axes. A \emph{density-precision axis} governs how many objects the model commits to and how token-level pressure is applied: a prompt-level output budget caps the requested number of objects, and a repetition penalty of 1.05 applies token-level pressure. A \emph{structural-integrity axis} governs whether the emitted list stays free of exact repeated records: object-level repeat-stop is a real generation-time stopping criterion that detects when an exact normalized object record would be emitted a second time and closes the kept prefix into a valid JSON array. Both axes probe the generation surface exposed by the fine-tuned model.

\paragraph{Metrics.} The metric set tracks parse-valid rate, mean predictions per image, class-aware one-to-one \fmetric, exact-object duplicate rate, maximum exact-object repeat, and repeat-stop trigger rate. F1 measures target signal; duplicate rate and max repeat expose repeated-tail pressure; parse-valid rate measures structure; trigger rate measures how often the structural control intervenes. We report image-level nonparametric bootstrap 95\% confidence intervals for \fmetric\ using 1000 resamples, and audit the promoted operating points with class-aware \fstrict.

\section{The High-Capacity Control Surface}

Table~\ref{tab:main} reports the main control-surface rows. The rows form the evidence chain: \modelgemma\ high-capacity raw adaptation exposes strong target signal plus repeated-tail pressure; repetition penalty shifts the surface; prompt budgeting selects a high-F1 density point; prompt-budget repeat-stop gives the clean high-capacity endpoint. \modelqwen\ reproduces the same density/repetition pattern and supplies a cross-family controlled endpoint.

\begin{table}[t]
\centering
\footnotesize
\setlength{\tabcolsep}{2.8pt}
\caption{Main control-surface evidence. Bracketed values are image-level bootstrap 95\% CIs for \fmetric; unbracketed \fmetric\ values and all control columns are point metrics, and dashes mark metrics not re-scored for that row. The \fstrict\ column audits the promoted operating points at the stricter IoU threshold.}
\label{tab:main}
\resizebox{\textwidth}{!}{%
\begin{tabular}{llrrrrrr}
\toprule
Model & Setting & Parse & Pred/img & \fmetric & \fstrict & Dup. & Max rep. \\
\midrule
\modelgemma & Base & 0.963 & 4.763 & 0.007 [0.000, 0.016] & 0.002 & 0.002 & 2 \\
\modelgemma & q/k/v/o r32 & 0.812 & 6.713 & 0.448 [0.384, 0.513] & 0.363 & 0.080 & 23 \\
\modelgemma & \quad + rep penalty 1.05 & 0.900 & 6.425 & 0.474 [0.408, 0.544] & 0.366 & 0.039 & 19 \\
\modelgemma & \quad + prompt budget 8 & 1.000 & 5.025 & 0.494 [0.428, 0.552] & 0.381 & 0.021 & 8 \\
\modelgemma & \quad + budget 8 + repeat-stop & 1.000 & 4.787 & 0.490 [0.427, 0.555] & 0.385 & 0.000 & 1 \\
\modelqwen & Base & 0.988 & 2.925 & 0.155 [0.118, 0.198] & 0.143 & 0.016 & 5 \\
\modelqwen & q/v r8 & 0.988 & 5.013 & 0.233 [0.163, 0.306] & 0.186 & 0.081 & 12 \\
\modelqwen & \quad + rep penalty 1.05 & 1.000 & 3.837 & 0.249 [0.189, 0.313] & 0.214 & 0.002 & 2 \\
\modelqwen & \quad + prompt budget 6 & 0.988 & 3.487 & 0.258 & -- & 0.044 & 6 \\
\modelqwen & \quad + budget 6 + repeat-stop & 1.000 & 3.475 & 0.259 & -- & 0.000 & 1 \\
\modelqwen & q/v r32 + budget 6 + stop & 1.000 & 4.838 & 0.318 [0.261, 0.384] & 0.275 & 0.000 & 1 \\
\bottomrule
\end{tabular}%
}
\end{table}

\begin{figure}[t]
\centering
\includegraphics[width=\textwidth]{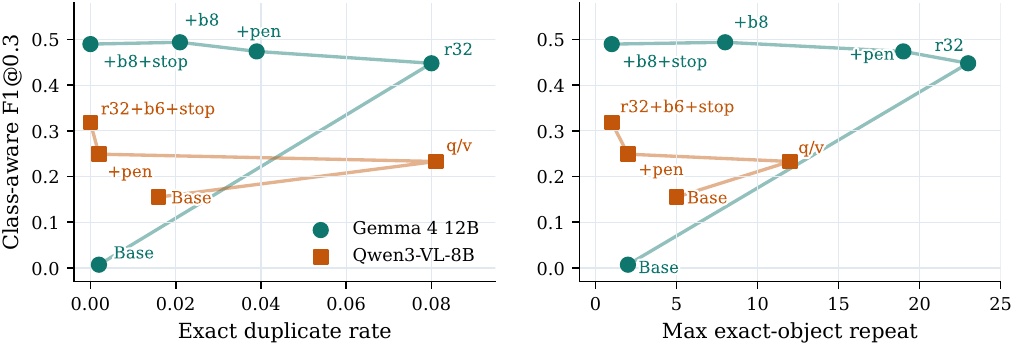}
\caption{Control-surface view of dense coordinate-list fine-tuning. Each labelled point is one operating configuration from Table~\ref{tab:main}. High-capacity Gemma adaptation exposes target signal and repeated-tail pressure on the same surface; Qwen reproduces the pattern at a different scale; the two control axes move both families toward the upper-left clean-termination region without losing target F1.}
\label{fig:control_surface}
\end{figure}

\modelgemma\ is the canonical surface. The base model is schema-stable on the JSON-bbox protocol (parse-valid rate 0.963, duplicate rate 0.002) but does not localize (\fmetric\ 0.007), so any structure the adapter later emits is induced by fine-tuning rather than inherited zero-shot. High-capacity q/k/v/o LoRA (rank 32) activates the target task strongly and, on the same generation surface, exposes repeated-tail pressure: the raw adapter reaches \fmetric\ 0.448 [0.384, 0.513] at 6.713 predictions per image with duplicate rate 0.080 and max repeat 23. Target localization and repeated-tail pressure rise together---the signature of dense coordinate-list interference---and we navigate this surface along the two orthogonal axes of Section~\ref{sec:setup}.

\begin{figure}[t]
\centering
\includegraphics[width=\textwidth]{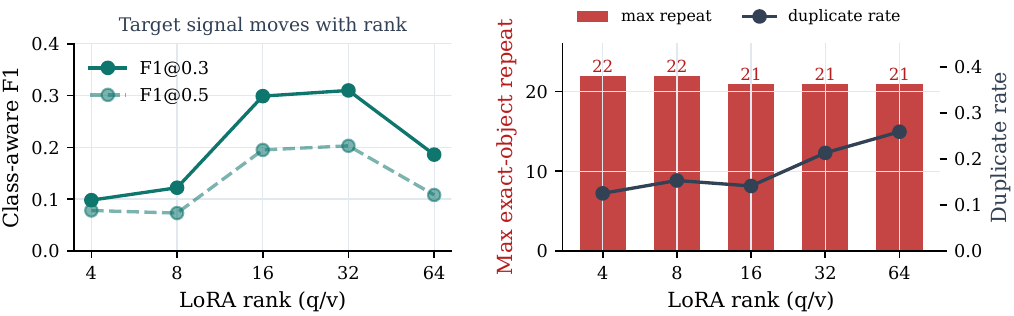}
\caption{Capacity persistence in the module-controlled q/v rank sweep. Target F1 (left) varies with rank and peaks at rank 32, while max exact-object repeat (right, bars) stays flat at 21--22 and duplicate rate (right, line) remains nonzero throughout: adapter capacity moves the target signal but does not remove the repeated tail.}
\label{fig:ranksweep}
\end{figure}

A module-controlled rank sweep confirms that this repeated-tail pressure is not removed by adapter capacity. Holding the Gemma q/v module set, training data, and 80-image evaluation split fixed, ranks 4/8/16/32/64 all retain exact repeated-tail pressure: max repeat stays in the narrow 21--22 band and duplicate rate remains nonzero (0.125--0.259). Figure~\ref{fig:ranksweep} plots the full sweep: target F1 moves with rank and peaks at rank 32, while the maximum-repeat band stays flat at 21--22 across the entire range. Rank changes target strength and duplicate frequency, but it does not erase the tail; the repeated-tail signature persists across more than an order of magnitude in trainable adapter parameters (2.6M at rank 4 to 41.5M at rank 64).

\paragraph{Density-precision axis.} Moving along this axis trades how many objects the model commits to against precision. A repetition penalty of 1.05 shifts the operating point to \fmetric\ 0.474 [0.408, 0.544] and lowers duplicate rate to 0.039, but max repeat stays at 19: token-level pressure improves the target point while the repeated tail persists. A prompt budget of eight items moves further to the strongest raw operating point, \fmetric\ 0.494 [0.428, 0.552] at parse-valid rate 1.000, with duplicate rate 0.021 and max repeat 8. Density tuning thus reaches a high-F1 point, but exact repeated records remain within the emitted list.

The density-precision axis is continuous, not a single setting. Sweeping the prompt budget from four to eight objects on the same 80-image evaluation split traces a controllable trajectory: \fmetric\ is 0.430 at budget four, 0.448 at five, 0.483 at six, 0.466 at seven, and 0.494 at eight, with parse-valid rate 1.000 at every budget (Table~\ref{tab:densitysweep}). Duplicate pressure stays low across the sweep (0.017--0.041) while maximum repeat tracks the budget ceiling, confirming that the prompt budget directly governs list density. Budget five reproduces the unbudgeted adapter's F1 (0.448) while lifting parse-valid rate from 0.812 to 1.000, separating format stability from the localization signal.

\begin{table}[t]
\centering
\footnotesize
\setlength{\tabcolsep}{3.5pt}
\caption{Density-precision axis sweep: high-capacity Gemma q/k/v/o r32 under prompt budgets four to eight on the 80-image evaluation split. Max repeat tracks the budget ceiling exactly.}
\label{tab:densitysweep}
\begin{tabular*}{\columnwidth}{@{\extracolsep{\fill}}rrrrrr}
\toprule
Budget & Pred/img & \fmetric & \fstrict & Dup. & Max rep. \\
\midrule
4 & 3.700 & 0.430 & 0.357 & 0.020 & 4 \\
5 & 4.150 & 0.448 & 0.364 & 0.041 & 5 \\
6 & 4.388 & 0.483 & 0.387 & 0.017 & 6 \\
7 & 4.750 & 0.466 & 0.364 & 0.021 & 7 \\
8 & 5.025 & 0.494 & 0.381 & 0.021 & 8 \\
\bottomrule
\end{tabular*}
\end{table}

The sweep makes the axis legible: density and target signal move together along a controllable curve rather than being fixed by the adapter. Budget eight is the strongest \fmetric\ density point (0.494; \fstrict\ peaks at budget six, 0.387), and applying repeat-stop on top of it reaches the clean endpoint. The two axes are complementary by construction---density-precision selects how many objects the model commits to, structural-integrity removes exact repeated records within that regime---so the high-F1 and clean-termination operating points coincide.

\begin{figure}[t]
\centering
\includegraphics[width=0.72\textwidth]{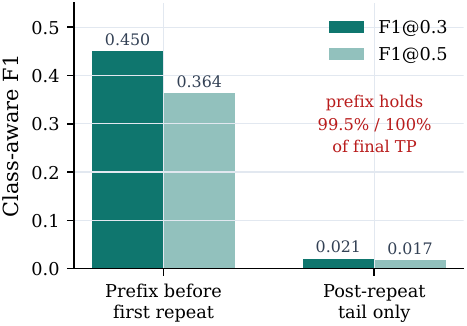}
\caption{Separability of the localization signal from the repeated tail. The prefix before the first exact repeat carries nearly all of the target F1 at both IoU thresholds; the post-repeat tail carries almost none.}
\label{fig:separability}
\end{figure}

\paragraph{Structural-integrity axis.} Object-level repeat-stop acts orthogonally: holding the budget-eight density point fixed, it closes the list at the first exact repeated record and reaches the clean high-capacity endpoint, \fmetric\ 0.490 [0.427, 0.555] at parse-valid rate 1.000 with duplicate rate 0.000 and max repeat 1, still emitting a dense 4.787 predictions per image. The trigger fires on only 6.2\% of decoding---removing the tail preserves target F1 (0.494 to 0.490, overlapping bootstrap CIs), while stricter \fstrict\ rises from 0.381 to 0.385 over the raw adapter's 0.363. A direct prefix analysis of the high-capacity raw outputs shows why: before the first exact repeated object, the prefix already captures 99.5\% of final class-matched TP at IoU 0.3 and 100.0\% at IoU 0.5, while the post-repeat tail alone has F1 0.021 at IoU 0.3 and 0.017 at IoU 0.5 (Figure~\ref{fig:separability}). The repeated tail is therefore a \emph{separable} low-value structural fault rather than a load-bearing part of the localization signal: the two axes reach the high-F1 point and the clean-termination point at the same place.

The two axes act on independent degrees of freedom, and a full budget $\times$ repeat-stop factorial makes this measurable. At every budget from four to eight, adding object-level repeat-stop forces duplicate rate to 0.000 and max repeat to 1 while moving \fmetric\ by at most $+0.010$/$-0.004$ (Table~\ref{tab:factorial}): repetition suppression and grounding strength are separate control dimensions at every density point, not only at the promoted endpoint.

\begin{table}[t]
\centering
\footnotesize
\setlength{\tabcolsep}{3.5pt}
\caption{Budget $\times$ repeat-stop factorial on the high-capacity Gemma adapter. Raw and +Stop columns report \fmetric; Dup., Max rep., and Trigger describe the +Stop condition.}
\label{tab:factorial}
\begin{tabular*}{\columnwidth}{@{\extracolsep{\fill}}rrrrrr}
\toprule
& & \multicolumn{4}{c}{+ object-level repeat-stop} \\
\cmidrule(lr){3-6}
Budget & Raw & \fmetric & Dup. & Max rep. & Trigger \\
\midrule
4 & 0.430 & 0.432 & 0.000 & 1 & 0.050 \\
5 & 0.448 & 0.458 & 0.000 & 1 & 0.075 \\
6 & 0.483 & 0.487 & 0.000 & 1 & 0.025 \\
7 & 0.466 & 0.473 & 0.000 & 1 & 0.025 \\
8 & 0.494 & 0.490 & 0.000 & 1 & 0.062 \\
\bottomrule
\end{tabular*}
\end{table}

The granularity of the interference predicts the failure mode of token-level control: the tail is made of exact repeated \emph{records}, so token-level constraints cannot remove it without breaking the format that carries it. The prediction holds (Table~\ref{tab:tokenbaseline}). Syntax-breaking controls fail outright---blocking repeated 3-grams destroys the coordinate format (parse-valid rate 0.287, \fmetric\ 0.000) because coordinate JSON is built from repeated token patterns, and 5-gram blocking collapses density (1.188 predictions per image, \fmetric\ 0.110). Threshold-sensitive controls fail structurally---a repetition penalty of 1.10 holds \fmetric\ at 0.492 but leaves max repeat at 19, and 1.20 suppresses the tail only at the cost of target signal (\fmetric\ 0.453), with no guarantee at either setting. Object-level repeat-stop is the matched intervention: it removes the tail by construction while holding the operating point, because it acts at the unit where the interference lives.

\begin{table}[t]
\centering
\footnotesize
\setlength{\tabcolsep}{2.5pt}
\caption{Token-level alternatives on the raw high-capacity Gemma adapter. N-gram blocking destroys the coordinate-list format; repetition penalty is threshold-sensitive with no structural guarantee; object-level repeat-stop removes the tail by construction.}
\label{tab:tokenbaseline}
\begin{tabular*}{\columnwidth}{@{\extracolsep{\fill}}lrrrrr}
\toprule
Decoding & Parse & Pred/img & \fmetric & Dup. & Max rep. \\
\midrule
raw & 0.812 & 6.713 & 0.448 & 0.080 & 23 \\
3-gram block & 0.287 & 0.237 & 0.000 & 0.000 & 1 \\
5-gram block & 0.575 & 1.188 & 0.110 & 0.000 & 1 \\
rep.\ pen.\ 1.10 & 0.963 & 5.763 & 0.492 & 0.020 & 19 \\
rep.\ pen.\ 1.20 & 0.975 & 5.912 & 0.453 & 0.000 & 1 \\
repeat-stop & 1.000 & 6.675 & 0.450 & 0.000 & 1 \\
\bottomrule
\end{tabular*}
\end{table}

\section{The Interference Is Bound to Coordinate-List Structure}

We now localize the pressure to its structural source. The repeated-tail pressure is specific to the bbox-coordinate object-list structure. Table~\ref{tab:structure} diagnoses this by holding the adapter and data fixed while varying only the output structure: every non-bbox row probes the \emph{same} bbox-trained adapter, unchanged, on a non-bbox task, so any pressure the adaptation injects would surface there. The q/v rows give a capacity-controlled structure sweep, isolating output structure under a matched rank-8 q/v adapter setting. The q/k/v/o rows then connect the non-bbox controls to the high-capacity Gemma setting in Table~\ref{tab:main}. The contrast is sharp in both model families: coordinate-list grounding exposes duplicate pressure, while equally structured non-bbox JSON outputs keep the repeated-record tail off the surface entirely.

\begin{table}[t]
\centering
\footnotesize
\setlength{\tabcolsep}{3.5pt}
\caption{Structure-axis specificity. Capacity-controlled rank-8 q/v rows isolate the structural trigger; high-capacity rank-32 q/k/v/o rows show that non-bbox JSON remains repeat-clean under the adapter used in Table~\ref{tab:main}, including the COCO-trained adapter. Target is class-aware \fmetric\ for bbox rows and target-shape validity rate for non-bbox JSON rows.}
\label{tab:structure}
\resizebox{\textwidth}{!}{%
\begin{tabular}{llrrrl}
\toprule
Model & Structure & Target & Dup. & Max rep. & Notes \\
\midrule
\modelgemma & bbox-coordinate list & 0.122 & 0.153 & 22 & repeated-tail pressure \\
\modelgemma & dense non-bbox JSON & 1.000 & 0.000 & 1 & rank-8 q/v \\
\modelgemma & spatial/count JSON & 1.000 & 0.000 & 1 & rank-8 q/v \\
\modelgemma & dense non-bbox JSON & 1.000 & 0.000 & 1 & rank-32 q/k/v/o \\
\modelgemma & spatial/count JSON & 1.000 & 0.000 & 1 & rank-32 q/k/v/o \\
\modelgemma & dense non-bbox JSON & 1.000 & 0.000 & 1 & rank-32 q/k/v/o (COCO) \\
\modelqwen & bbox-coordinate list & 0.233 & 0.081 & 12 & repeated-tail pressure \\
\modelqwen & dense non-bbox JSON & 1.000 & 0.000 & 1 & rank-8 q/v \\
\modelqwen & spatial/count JSON & 1.000 & 0.000 & 1 & rank-8 q/v \\
\bottomrule
\end{tabular}%
}
\end{table}

This diagnosis localizes the mechanism. The pressure concentrates in the coordinate-list regime where visual localization, numeric serialization, repeated object classes, and termination interact, and nowhere else: non-bbox JSON stays repeat-clean under both the capacity-controlled rank-8 q/v adapter and the high-capacity q/k/v/o r32 adapter, while bbox-coordinate lists remain the high-pressure surface. Having localized the pressure to coordinate-list structure in both model families, we next confirm that the mechanism and its two-axis controllability are not Gemma-specific.

\section{The Surface Is Cross-Family: Qwen Reproduction}

\modelqwen\ reproduces the interference signature under a different architecture and coordinate protocol: q/v adaptation raises \fmetric\ from 0.155 [0.118, 0.198] to 0.233 [0.163, 0.306] while duplicate rate rises from 0.016 to 0.081 and max repeat from 5 to 12 (Table~\ref{tab:main})---stronger target signal and repeated-tail pressure move together on one surface.

The two axes navigate this surface in the same directions. On the density-precision axis, repetition penalty 1.05 shifts the operating point (\fmetric\ 0.249 [0.189, 0.313], max repeat 2) and a budget of six selects the density region; on the structural-integrity axis, repeat-stop holds \fmetric\ (0.258 to 0.259) while forcing duplicate rate to 0.000 and max repeat to 1 at a trigger rate of 5.0\%. Capacity persistence transfers as well: target signal rises with rank while the repeated tail persists across ranks 8--64 (duplicate rate 0.006--0.044, max repeat 3--6). The high-capacity rank-32 endpoint completes the reproduction: parse-valid rate 1.000, \fmetric\ 0.318 [0.261, 0.384], \fstrict\ 0.275, duplicate rate 0.000, max repeat 1, trigger rate 2.5\%---the same clean termination reached on Gemma.

Severity and best operating points differ by family; the surface shape and its two-axis response do not.

\section{The Surface Generalizes: COCO 2017 Reproduction}

We next confirm that the coordinate-list acquisition and repetition signature transfers beyond the original industrial dataset. COCO 2017 supplies public dense detection annotations with varied categories, clutter, and co-occurring objects. We use the object-centric split described in Section~\ref{sec:setup} and evaluate base and adapted \modelgemma\ under the same prompt budget of six, matching the maximum target cardinality of the evaluation set.

\begin{table}[t]
\centering
\footnotesize
\setlength{\tabcolsep}{3.5pt}
\caption{Cross-dataset reproduction on a public COCO 2017 object-centric subset, at two data scales. Recall columns are image-level mean class-aware recall; F1 columns are micro class-aware one-to-one F1. Adapted rows use the same high-capacity q/k/v/o rank-32 setup as the main Gemma chain; dashes mark metrics not re-scored for that row.}
\label{tab:coco}
\resizebox{\textwidth}{!}{%
\begin{tabular}{lrrrrrrrrr}
\toprule
Condition & Parse & Pred/img & Max IoU & Rec@0.3 & Rec@0.5 & F1@0.3 & F1@0.5 & Dup. & Max rep. \\
\midrule
\multicolumn{10}{@{}l}{\textit{780 train / 120 eval}} \\
Base + budget 6 & 1.000 & 1.358 & 0.060 & 0.0228 & 0.0128 & 0.0282 & 0.0141 & 0.000 & 1 \\
q/k/v/o r32 + budget 6 & 1.000 & 2.192 & 0.234 & 0.1540 & 0.0981 & 0.1647 & 0.1048 & 0.016 & 4 \\
\quad + repeat-stop & 1.000 & 2.108 & 0.234 & 0.1540 & -- & 0.1672 & 0.1064 & 0.000 & 1 \\
\midrule
\multicolumn{10}{@{}l}{\textit{160 train / 80 eval (InsPLAD-scale)}} \\
Base + budget 6 & 1.000 & 1.100 & 0.063 & 0.0104 & 0.0021 & 0.0180 & 0.0060 & 0.000 & 1 \\
q/k/v/o r32 + budget 6 & 1.000 & 2.550 & 0.519 & 0.4315 & 0.3575 & 0.4321 & 0.3697 & 0.009 & 2 \\
\bottomrule
\end{tabular}%
}
\end{table}

\begin{figure}[t]
\centering
\includegraphics[width=0.74\textwidth]{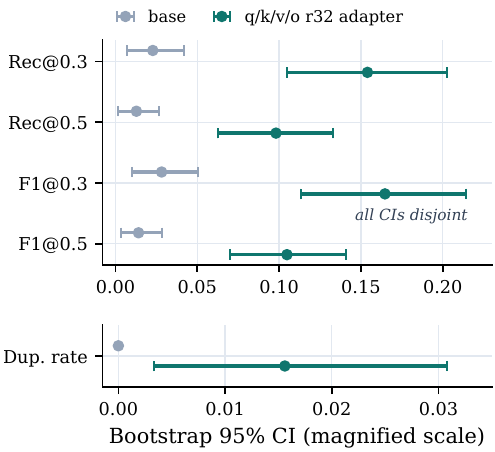}
\caption{Image-level bootstrap 95\% CIs for the COCO reproduction (120-image evaluation split, 1000 resamples). Base and adapted CIs are disjoint on every target metric, and the adapted duplicate-rate CI excludes zero: the acquired localization and the transferred duplicate pressure are both statistically resolved.}
\label{fig:cococi}
\end{figure}

Table~\ref{tab:coco} shows a clean transfer of the dense coordinate-list surface. The base model is parse-stable under the COCO schema but has low localization under the dense-bbox protocol: class-aware recall@0.3 is 0.0228 and recall@0.5 is 0.0128. The same high-capacity q/k/v/o adaptation raises recall@0.3 to 0.1540 and recall@0.5 to 0.0981, a 6.76$\times$ and 7.67$\times$ lift respectively, while mean max IoU rises from 0.060 to 0.234. The repeated-list signature transfers as well: exact duplicate rate rises from 0.000 to 0.016 and max repeat rises from 1 to 4. Image-level bootstrap CIs separate the two rows on every target metric (Figure~\ref{fig:cococi}), and the adapted duplicate-rate CI [0.0033, 0.0308] excludes zero, so the acquired duplicate pressure is statistically significant rather than incidental. The structural-integrity axis closes the loop on this dataset as well: adding object-level repeat-stop to the same operating point removes the transferred tail (duplicate rate 0.016 to 0.000, max repeat 4 to 1) at a trigger rate of 4.2\%, while \fmetric\ rises from 0.1647 to 0.1672---the full two-axis mechanism, not only the interference signature, reproduces on COCO.

Training-set scale does not drive the surface (Table~\ref{tab:coco}, bottom block). Training on 160 images and evaluating on 80 images from the same COCO subset, the adapter reaches F1@0.3 0.4321 and F1@0.5 0.3697 against a matched base of 0.0180 and 0.0060, with mean max IoU rising from 0.063 to 0.519 and duplicate pressure again emerging (duplicate rate 0.000 to 0.009, max repeat 1 to 2). This second-dataset result confirms that the coordinate-list interference surface generalizes to public dense detection data at both data scales.

\section{Reproducibility Details}

\paragraph{Training.} All adapters are standard LoRA fine-tunes on the frozen base models. The Gemma q/v rank-8 adapter trains for 40 optimization steps on the 160-image split with learning rate $10^{-4}$ and $\alpha=16$; the rank-sweep adapters (ranks 4/16/32/64) follow the same recipe with $\alpha$ fixed at twice the rank. The high-capacity Gemma q/k/v/o rank-32 adapter ($\alpha=64$, 42.7M trainable parameters) trains for 120 steps on the same 160 images. The COCO adapters use the identical q/k/v/o rank-32 configuration on the COCO training splits described in Section~\ref{sec:setup}.

\paragraph{Decoding and controls.} Bbox evaluation decodes up to 768 new tokens; dense non-bbox JSON probes use 1024. The dense non-bbox probe requests a fixed eight-slot inventory with fixed keys, and its outputs average 2.3k characters---several times the typical bbox-list output. The prompt budget is a natural-language output constraint (``return at most $N$ objects'') applied identically to base and adapted models. The repetition penalty, where stated, is 1.05. Object-level repeat-stop normalizes each emitted object record (class string and coordinate values) during generation and closes the JSON array when an exact normalized record would be emitted a second time; the trigger rate reports how often this criterion fires.

\paragraph{Evaluation.} Class-aware one-to-one F1 matches predictions to ground truth greedily by IoU within each class, at thresholds 0.3 and 0.5. Exact-object duplicate rate and max repeat are computed over normalized records per image.

\section{Relation to Prior Work}

\paragraph{Interference and forgetting under adaptation.} Multimodal instruction tuning and continual tuning work shows that large multimodal models can exhibit task interference, response-format drift, and forgetting under adaptation \cite{continual-lmm,smolora,vision-flan}, and PEFT studies show that adapter choices affect stability, generalization, and hallucination behavior in MLLMs \cite{mixlora,peft-mllm}. This line typically frames the effect as broad capability loss. We instead localize the effect: a high-capacity adapter that raises target grounding leaves dense non-bbox JSON and spatial/count generation fully intact, and the measurable pressure concentrates on bbox-coordinate object lists.

\paragraph{Visual grounding.} Grounding work studies how multimodal models localize objects and how design choices affect target performance \cite{expvg,few-heads-grounding}. Coordinate-as-text grounding, in which the model emits box coordinates as ordinary tokens \cite{kosmos2,shikra}, makes dense localization a structured-generation problem---precisely the regime our surface analysis targets. We use grounding as the behavioral metric that exposes the generation surface: the dense coordinate list is the instrument, and its repeated-tail behavior is the signal of interest.

\paragraph{Inference-time control.} Decoding-time control work shows that object-level generation behavior can be shaped at inference time \cite{clip-guided-decoding,woodpecker}. Object-hallucination metrics ask whether described objects exist in the image \cite{chair,pope}; the repeated tail is a different failure mode---records that are visually grounded but structurally redundant. We use repetition penalty, prompt budgeting, and object-level repeat-stop as probes that separate useful localization from exact repeated records, mapping the control surface exposed by the fine-tuned model.

\paragraph{Text degeneration and constrained decoding.} Neural text generation is known to fall into repetition under certain decoding regimes \cite{holtzman-degeneration}, and training-time objectives have been proposed to penalize repeated continuations \cite{welleck-unlikelihood,ditto}; constrained-decoding frameworks enforce output schemas at the token level \cite{willard-outlines}. The repeated tail we measure differs from generic text degeneration on three measured grounds: it is induced by fine-tuning rather than by the decoding regime (the base model shows near-zero duplicate pressure, 0.002, on the same protocol), it is specific to bbox-coordinate lists rather than to long structured output in general (longer dense non-bbox JSON outputs stay clean), and it operates at the object-record level rather than the token level, which is why an object-level stopping criterion removes it cleanly while token-level repetition penalty leaves max repeat at 19.

The dense coordinate-list surface connects these threads into a localized fine-tuning interference mechanism: dense bbox adaptation creates target signal and repeated-tail pressure on the same generation surface, while lightweight controls separate useful localization from exact repeated records.

\section{Conclusion}

The coordinate-list surface yields a design rule: an adapter that learns to localize will also learn to repeat, and the two behaviors occupy separable regions of the same generation surface. Dense grounding evaluation that reports only target F1 therefore measures half of what fine-tuning changes---duplicate rate and max repeat belong next to F1 in any dense structured-output comparison.

The two-axis decomposition is directly actionable. A practitioner fine-tuning a multimodal LM for dense grounding can select density with a prompt budget, audit the surface with the paired repetition metrics, and deploy object-level repeat-stop to guarantee duplicate-free, validly terminated lists---without retraining and without giving up the localization gain, on either model family and on both datasets.

The granularity result constrains control design in the same way: token-level repetition machinery is structurally mismatched with record-level interference, so structured-output regimes need controls at the unit of structure. The interference itself is structure-specific, capacity-persistent, cross-family, and separable from the target signal, and it reproduces on a second public dataset: a localized, measurable, and decomposable generation surface rather than diffuse capability loss.

\end{document}